\documentclass[3p,review,authoryear]{elsarticle}
\DeclareUnicodeCharacter{0308}{\forall}
\usepackage{geometry}                                   \usepackage{graphicx}
\usepackage{natbib}
\usepackage{enumitem}
\usepackage{adjustbox}
\usepackage{amssymb}
\usepackage{rotating}
\usepackage{hyperref}
\usepackage{booktabs}
\usepackage{subfig}
\usepackage{engord}
\usepackage{url}
\usepackage{mathtools}
\usepackage{amsmath}

\usepackage{todonotes}
\usepackage[normalem]{ulem}

\usepackage[nomargin,inline,marginclue,draft]{fixme}
 \fxsetup{theme=color, mode=multiuser} 
 \FXRegisterAuthor{pb}{apb}{\color{blue}PB} 
 \FXRegisterAuthor{va}{avs}{\color{red}VA} 
 \FXRegisterAuthor{bmb}{abmb}{\color{blue}BMB}

\usepackage{todonotes}
\usepackage{xargs} 
\usepackage{tabularx}
\newcommandx{\pbtodo}[2][1=]{\todo[linecolor=red,backgroundcolor=red!25,bordercolor=red,#1]{\tiny PB:  #2}}

\newcommandx{\vatodo}[2][1=]{\todo[linecolor=blue,backgroundcolor=blue!25,bordercolor=blue,#1]{\tiny VA: #2}}

\newcommandx{\bmbtodo}[2][1=]{\todo[linecolor=blue,backgroundcolor=blue!55,bordercolor=blue,#1]{\tiny BMB: #2}}

\usepackage[commandnameprefix=ifneeded]{changes}

\setcommentmarkup{\linepenalty=1000\todo[color=authorcolor!10, bordercolor=authorcolor, linecolor=authorcolor!70, nolist]{\textbf{[#3\penalty1000]} \hskip 0pt plus1fill\relax\penalty50\hskip 0pt plus-1fill\relax#1}}

\definechangesauthor[name=r1, color=brown]{r1}
\definechangesauthor[name=r2, color=orange]{r2}

\usepackage{verbatim}
\usepackage{arydshln}

\usepackage{algorithm}
\usepackage{algpseudocode}
\usepackage{pythontex}
\usepackage{graphicx}

\usepackage{tikz}
\usetikzlibrary{shapes.geometric, arrows.meta, positioning, calc, patterns, decorations.pathreplacing}

\usepackage[acronym]{glossaries}
\newacronym{ml}{ML}{Machine Learning}
\newacronym{spyct}{SPYCT}{oblique predictive clustering tree}
\newacronym{vspyct}{VSPYCT}{variational oblique predictive clustering tree}
\newacronym{opct}{OPCT}{option predictive clustering tree}
\newacronym{sslmtr}{SSL-MTR}{semi-supervised multi-target regression}
\newacronym{pct}{PCT}{predictive clustering tree}
\newacronym{sop}{SOP}{structured output prediction}
\newacronym{fp}{FP}{False Positive}
\newacronym{fn}{FN}{False Negative}
\newacronym{tpr}{TPR}{True Positive Rate}
\newacronym{fpr}{FPR}{False Positive Rate}
\newacronym{auc}{AUC}{Area under the ROC curve}
\newacronym{tp}{TP}{True Positives}
\newacronym{tn}{TN}{True Negatives}
\newacronym{ecdf}{ECDF}{Empirical Cumulative Distribution Function}
\newacronym{isce}{ISCED}{International Standard Classification of Education}
\newacronym{kl}{KL}{Kullback–Leibler}
\newacronym{elbo}{ELBO}{Evidence Lower Bound}
\newacronym{gp}{GP}{Gaussian Process}

\graphicspath{ {./}{../figures/}{../../experiments/} }

\begin{document}
\begin{frontmatter}
	\title{Uncertainty-Aware Extrapolation in Bayesian Oblique Trees}
	\author[1,2]{Viktor Andonovikj\corref{cor1}}
	\ead{viktor.andonovikj@ijs.si}
	\author[1]{Sašo Džeroski}
	\ead{saso.dzeroski@ijs.si}
	\author[1,3]{Pavle Boškoski}
	\ead{pavle.boskoski@ijs.si}
	\address[1]{Jožef Stefan Institute, Jamova cesta 39, 1000 Ljubljana, Slovenia}
	\address[2]{Jožef Stefan International Postgraduate School, Jamova cesta 39, 1000 Ljubljana, Slovenia}
	\address[3]{Faculty of Information Studies in Novo mesto, Ljubljanska cesta 31b, 8000 Novo mesto, Slovenia}
	\cortext[cor1]{Corresponding author.} 
\begin{abstract}

Decision trees are widely used due to their interpretability and efficiency, but they struggle in regression tasks that require reliable extrapolation and well-calibrated uncertainty.
Piecewise-constant leaf predictions are bounded by the training targets and often become overconfident under distribution shift.
We propose a single-tree Bayesian model that extends \glspl{vspyct} by equipping each leaf with a \gls{gp} predictor.
Bayesian oblique splits provide uncertainty-aware partitioning of the input space, while \gls{gp} leaves model local functional behaviour and enable principled extrapolation beyond the observed target range.
We present an efficient inference and prediction scheme that combines posterior sampling of split parameters with \gls{gp} posterior predictions, and a gating mechanism that activates \gls{gp}-based extrapolation when inputs fall outside the training support of a leaf.
Experiments on benchmark regression tasks show improvements in the predictive performance compared to standard variational oblique trees, and substantial performance gains in extrapolation scenarios.

\end{abstract}
\end{frontmatter}
\glsresetall

\section{Introduction}

Decision trees and tree-based models play a central role in \gls{ml} due to their interpretability, modular structure, and strong empirical performance on many tasks~\cite{Mienye_2024}.
Within the \gls{sop} paradigm, \glspl{pct} have emerged as a unifying framework for predicting structured output in different settings, such as regression, classification, multi-target prediction.
Additional benefit is that they can work in a semi-supervised setting, which enables them to leverage large samples of unlabeled data~\cite{Levati__2018}.

To overcome the limitations of axis-aligned splits, \glspl{spyct}~\citep{Stepi_nik_2021} were introduced, allowing splits defined by linear combinations of input features.
Oblique splits enable more expressive decision boundaries, particularly in high-dimensional and sparse settings, and have been shown to improve predictive performance across a wide range of \gls{sop} tasks~\citep{Stepi_nik_2021_semi}.
However, \glspl{spyct} typically rely on ensembles to achieve state-of-the-art accuracy, which weakens interpretability.

Recently, \glspl{vspyct} addressed this limitation by introducing a Bayesian treatment of split parameters~\citep{Andonovikj_2026}.
By modelling the oblique split parameters as random variables and learning their posterior distributions using variational inference~\citep{Blei_2017}, \glspl{vspyct} integrate uncertainty directly into the decision process of a single tree.
This allows competitive performance relative to ensembles while preserving interpretability and enabling uncertainty quantification.
\Glspl{vspyct} thus represent an important step toward fully Bayesian, interpretable models for \gls{sop}.

Despite these advances, an important limitation remains.
Like most tree-based regression models, \glspl{vspyct} produce constant predictions at the leaves, corresponding to prototype values derived from training data.
As a consequence, predictions are inherently bounded by the observed target range in the training data.
This prevents meaningful extrapolation and often leads to overconfident predictions when inputs fall outside the training support.
The issue is particularly relevant in \gls{sop} settings involving temporal data~\citep{Bontempi_2013}, survival analysis~\citep{Andonovikj_2024, Ishwaran_2008}, economic indicators, or any application where future or extreme outcomes must be anticipated rather than interpolated.

This limitation highlights a missing component in the current \gls{vspyct} framework: while the uncertainty in the splits is modelled explicitly, the uncertainty in functional behaviour within leaves is not.
In other words, \gls{vspyct} captures where an instance should be routed, but not how the target should behave beyond observed data once a region of the input space has been selected.
Addressing this gap is the main goal of this paper and it is essential for extending Bayesian \glspl{pct} to extrapolation-sensitive tasks.

\begin{figure}[!htb]
\centering
\begin{tikzpicture}[
trainpt/.style={circle, fill=blue!70, inner sep=0pt, minimum size=5pt},
    ]

\begin{scope}
\draw[-{Stealth}, gray!70, thick] (-0.5, 0) -- (12, 0) node[right, font=\small] {$x$};
        \draw[-{Stealth}, gray!70, thick] (0, -0.5) -- (0, 5.5) node[above, font=\small] {$y$};

\fill[blue!8] (2.5, 0) rectangle (8.5, 5.2);

\draw[blue!40, dashed, thick] (2.5, 0) -- (2.5, 5.2);
        \draw[blue!40, dashed, thick] (8.5, 0) -- (8.5, 5.2);

\node[font=\scriptsize, text=orange!70!red] at (1.2, 4.8) {Extrapolation};
        \node[font=\scriptsize, text=blue!60] at (5.5, 4.8) {Training region};
        \node[font=\scriptsize, text=orange!70!red] at (10.2, 4.8) {Extrapolation};

\draw[gray!80, thick, dashed]
            (0.5, 1.2) .. controls (2, 1.6) and (4, 2.2) .. (5.5, 2.8)
            .. controls (7, 3.4) and (9, 4.0) .. (11.5, 5.0);
        \node[font=\scriptsize, text=gray!90, right] at (11.7, 5.0) {$f(x)$};

\foreach \x/\y in {2.8/1.9, 3.2/2.0, 3.6/2.15, 4.0/2.25, 4.4/2.35,
                          4.8/2.5, 5.2/2.65, 5.6/2.8, 6.0/2.9, 6.4/3.05,
                          6.8/3.15, 7.2/3.3, 7.6/3.45, 8.0/3.55, 8.3/3.65} {
            \node[trainpt] at (\x, \y) {};
        }

\draw[violet!80!black, very thick]
            (2.5, 1.90) .. controls (4, 2.20) and (5.5, 2.70) .. (5.5, 2.75)
            .. controls (5.5, 2.80) and (7, 3.30) .. (8.5, 3.60);

\draw[teal!80!black, very thick]
            (2.5, 2.00) .. controls (4, 2.30) and (5.5, 2.80) .. (5.5, 2.85)
            .. controls (5.5, 2.90) and (7, 3.40) .. (8.5, 3.70);

\draw[violet!80!black, very thick]
            (0.5, 1.90) -- (1.8, 1.90)
            .. controls (2.1, 1.90) and (2.3, 1.90) .. (2.5, 1.90);

\draw[violet!80!black, very thick]
            (8.5, 3.60) .. controls (8.7, 3.62) and (8.9, 3.63) .. (9.2, 3.63)
            -- (11.5, 3.63);

\fill[violet!15, opacity=0.5]
            (0.5, 1.75) rectangle (2.5, 2.05);
        \fill[violet!15, opacity=0.5]
            (2.5, 1.70) .. controls (4, 2.00) and (5.5, 2.50) .. (5.5, 2.55)
            .. controls (5.5, 2.60) and (7, 3.10) .. (8.5, 3.40)
            -- (8.5, 3.80)
            .. controls (7, 3.50) and (5.5, 2.95) .. (5.5, 2.95)
            .. controls (5.5, 2.90) and (4, 2.40) .. (2.5, 2.10)
            -- cycle;
        \fill[violet!15, opacity=0.5]
            (8.5, 3.45) rectangle (11.5, 3.81);

\draw[teal!80!black, very thick]
            (0.5, 1.20) .. controls (1.5, 1.55) and (2.0, 1.80) .. (2.5, 2.00);

\draw[teal!80!black, very thick]
            (8.5, 3.70) .. controls (9.5, 4.05) and (10.5, 4.55) .. (11.5, 5.15);

\fill[teal!20, opacity=0.7]
            (0.5, 0.35) .. controls (1.5, 0.85) and (2.0, 1.40) .. (2.5, 1.80)
            -- (2.5, 2.20)
            .. controls (2.0, 2.20) and (1.5, 2.05) .. (0.5, 2.05)
            -- cycle;

\fill[teal!15, opacity=0.4]
            (2.5, 1.80) .. controls (4, 2.10) and (5.5, 2.60) .. (5.5, 2.65)
            .. controls (5.5, 2.70) and (7, 3.20) .. (8.5, 3.50)
            -- (8.5, 3.90)
            .. controls (7, 3.60) and (5.5, 3.05) .. (5.5, 3.05)
            .. controls (5.5, 3.00) and (4, 2.50) .. (2.5, 2.20)
            -- cycle;

\fill[teal!20, opacity=0.7]
            (8.5, 3.50) .. controls (9.5, 3.75) and (10.5, 3.85) .. (11.5, 3.85)
            -- (11.5, 6.45)
            .. controls (10.5, 5.25) and (9.5, 4.35) .. (8.5, 3.90)
            -- cycle;

\begin{scope}[yshift=-1.5cm, xshift=1.5cm]
\node[trainpt] at (0, 0) {};
            \node[font=\scriptsize, right] at (0.2, 0) {Training data};

\draw[gray!80, thick, dashed] (2.3, 0) -- (3.1, 0);
            \node[font=\scriptsize, right] at (3.3, 0) {True $f(x)$};

\draw[violet!80!black, very thick] (5.3, 0.05) -- (6.1, 0.05);
            \fill[violet!15, opacity=0.6] (5.3, -0.1) rectangle (6.1, 0.2);
            \node[font=\scriptsize, right] at (6.3, 0.05) {VSPYCT};

\draw[teal!80!black, very thick] (8.5, 0) .. controls (8.8, 0.1) .. (9.3, 0.2);
            \fill[teal!20, opacity=0.7] (8.5, -0.15) rectangle (9.3, 0.35);
            \node[font=\scriptsize, right] at (9.5, 0.05) {VSPYCT-GP};
        \end{scope}

\draw[-{Stealth}, gray!60, thin] (10.3, 2.9) -- (10.3, 3.50);
        \node[font=\tiny, text=violet!70!black, align=center] at (10.3, 2.5) {Bounded\\(flat)};

\draw[{Stealth}-{Stealth}, teal!60, thin] (11.3, 3.95) -- (11.3, 6.25);
        \node[font=\tiny, text=teal!60, rotate=90] at (11.7, 5.0) {Growing uncertainty};

\node[font=\tiny, text=black!60, align=center] at (5.5, 4.2) {Similar predictions\\in training region};
        \draw[-{Stealth}, gray!40, thin] (5.5, 3.9) -- (5.5, 3.1);

    \end{scope}

\end{tikzpicture}
\caption{Extrapolation behaviour on synthetic 1D regression data.
The true function $f(x)$ (dashed gray) exhibits a smooth upward trend.
Training data (blue dots) are observed only within a bounded region.
\textbf{In the training region}, both VSPYCT (violet) and VSPYCT-GP (teal) produce similar smooth predictions due to Monte Carlo averaging over probabilistic split parameters.
\textbf{In extrapolation regions}, VSPYCT predictions flatten---bounded by the range of leaf prototypes---while VSPYCT-GP continues the learned trend with confidence intervals that widen to reflect increasing uncertainty.
This illustrates how GP leaf models enable calibrated extrapolation while preserving in-distribution performance.}
\label{fig:extrapolation-intuition}
\end{figure}
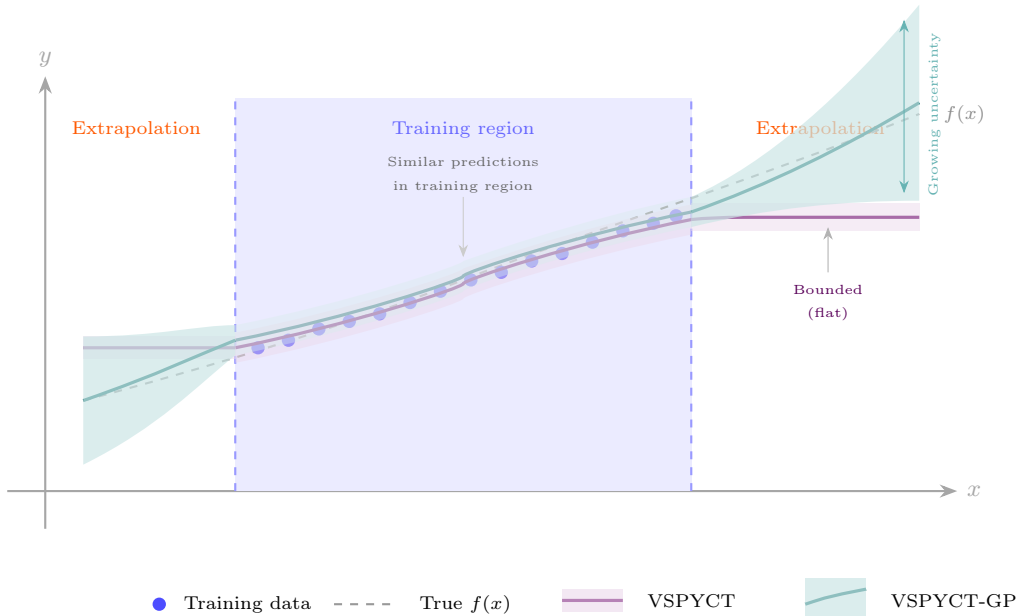

We propose an extension that equips \gls{vspyct} leaves with \gls{gp} predictors.
\Glspl{gp} provide a Bayesian, nonparametric model over functions, naturally combining interpolation, extrapolation, and calibrated uncertainty.
By integrating \gls{gp} leaves into a variational oblique tree, we decouple partitioning uncertainty from functional uncertainty: oblique Bayesian splits determine where an instance should be routed, while \gls{gp} leaves determine how the target behaves within and beyond the observed region from the training data.
To preserve strong in-distribution performance, we introduce a gating mechanism that activates \gls{gp}-driven extrapolation when inputs fall outside the training support of a leaf (see Figure~\ref{fig:extrapolation-intuition}).

The main contributions of this paper are as follows:
\begin{enumerate}[label=(\alph*)]

\item Model extension: We introduce a hybrid Bayesian tree model that combines \gls{vspyct} with \gls{gp} leaf models, enabling uncertainty-aware functional prediction within each leaf.

\item Principled extrapolation: We show how \gls{gp} leaves allow calibrated extrapolation beyond the training target range, addressing a fundamental limitation of \gls{vspyct} and tree-based predictors more broadly.

\item Inference and prediction: We develop an efficient inference and prediction procedure that integrates posterior sampling of oblique split parameters with closed-form \gls{gp} posterior predictions, preserving the single-tree structure and interpretability.

\end{enumerate}

We demonstrate improved tail-region predictive power, and better uncertainty calibration under controlled extrapolation scenarios, while maintaining on-par, and in many cases improved in-distribution performance compared to \glspl{vspyct}.
The proposed extension preserves the interpretability advantages of \glspl{vspyct} and enables additional performance gains in extrapolation scenarios.

\section{Related work}

Numerous extensions have been proposed to improve the predictive performance of \glspl{pct}, most notably through ensemble methods and more expressive split functions~\citep{Stepisnik_2020, Kocev_2013}.
\Glspl{spyct}~\citep{Stepi_nik_2021} replace axis-aligned splits with linear combinations of input features, enabling more flexible partitions of the input space and improved performance in high-dimensional settings.
However, like most tree-based models, both \glspl{pct} and \glspl{spyct} rely on constant predictions at the leaves, which limits their ability to extrapolate beyond the range of observed target values.

Recent work on \glspl{vspyct}~\citep{Andonovikj_2026} introduced a Bayesian treatment of oblique split parameters, allowing uncertainty to be modelled directly at the decision level within a single tree.
While this approach improves robustness and uncertainty estimation in the routing process, predictions at the leaf level remain bounded by training targets, and extrapolation behaviour is unchanged.
Addressing this limitation is particularly important in \gls{sop} scenarios where future, extreme, or out-of-distribution outcomes must be predicted.

Model trees extend regression trees by fitting parametric models, typically linear regressors, in the leaves instead of constant prototypes~\citep{Frank_1998}.
Classical approaches such as M5 and its variants demonstrate that piecewise linear modelling can improve predictive accuracy by capturing local trends.
These methods also enable limited extrapolation within each leaf, as predictions are no longer constant.
However, model trees require a fixed functional form to be chosen a priori and generally provide no principled mechanism for uncertainty quantification.
In practice, linear leaf models can extrapolate aggressively and with unjustified confidence~\citep{Loh_2007, Raymaekers_2024}, especially when leaves contain few or biased observations.
As a result, while model trees partially alleviate the extrapolation problem, they do not address uncertainty-aware extrapolation in a Bayesian sense.

An alternative line of work combines tree-based partitioning with \gls{gp} regression.
Treed Gaussian Process models partition the input space using a tree structure and fit a separate GP in each region, enabling local non-stationary modelling and principled uncertainty estimation.
More recent approaches integrate \glspl{gp} with Bayesian additive regression trees~\citep{Chipman_2010}, showing improved extrapolation and calibration compared to standard tree ensembles.
These models demonstrate the effectiveness of combining partitioning with nonparametric Bayesian regression, particularly for extrapolation-sensitive tasks.

Despite their strengths, treed \gls{gp} models typically rely on axis-aligned splits, complex MCMC-based inference, or ensemble formulations, which limits scalability and interpretability.
Moreover, they are not designed for the \gls{sop} setting addressed by predictive clustering trees, nor do they integrate naturally with oblique split structures learned via variational inference.

The present work builds directly on \gls{vspyct} by addressing a complementary source of uncertainty: functional uncertainty within leaves. Instead of replacing the tree structure or resorting to ensembles, we augment \gls{vspyct} with Gaussian process leaf models, enabling local, uncertainty-aware functional prediction while preserving the oblique Bayesian partitioning learned via variational inference~\citep{hoffman2013stochastic}.
In contrast to model trees, this approach avoids committing to a fixed parametric form, and unlike treed \gls{gp} or BART-based methods, it maintains a single-tree structure tailored to structured output prediction.
As a result, the proposed model directly targets the extrapolation limitations of existing predictive clustering trees while remaining consistent with the probabilistic and algorithmic design of \gls{vspyct}.

\section{The VSPYCT-GP Model}
\label{sec:vspyct-gp}

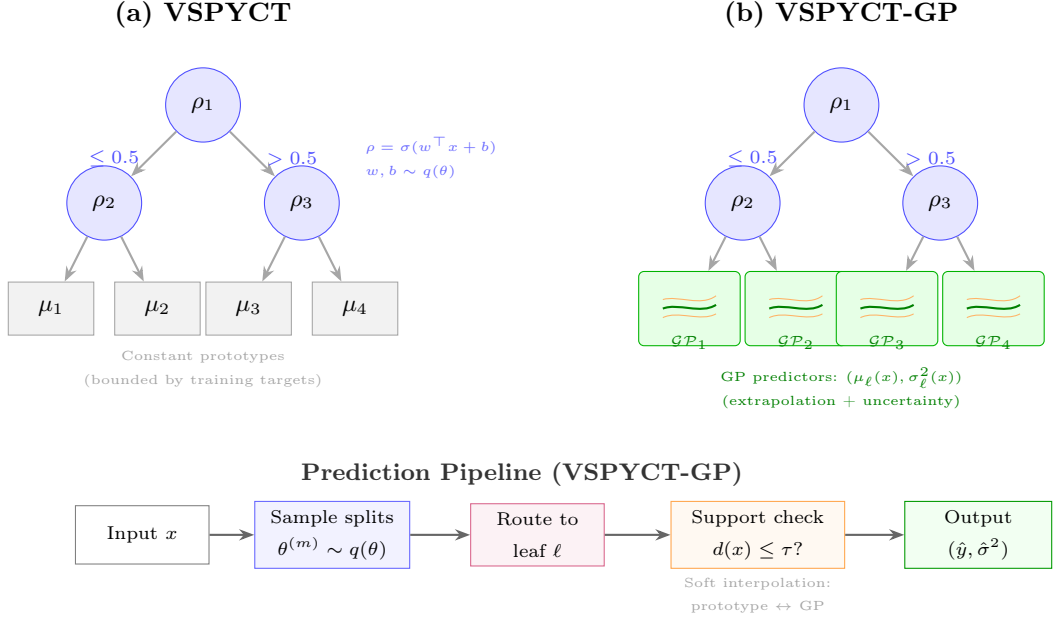
\begin{figure}[!htb]
\centering
\begin{tikzpicture}[
split/.style={circle, draw=blue!70, fill=blue!10, minimum size=28pt, font=\small},
    leaf/.style={rectangle, draw=gray!70, fill=gray!10, minimum width=32pt, minimum height=20pt, font=\small},
    gpleaf/.style={rectangle, draw=green!70!black, fill=green!10, minimum width=38pt, minimum height=28pt, font=\small, rounded corners=2pt},
edge/.style={-{Stealth[length=2mm]}, thick, gray!70},
splitlabel/.style={font=\scriptsize, text=blue!70},
    ]

\begin{scope}[xshift=-4.2cm]
\node[font=\bfseries] at (0, 2.2) {(a) VSPYCT};

\node[split] (root1) at (0, 1) {$\rho_1$};

\node[split] (n2) at (-1.3, -0.3) {$\rho_2$};
        \node[split] (n3) at (1.3, -0.3) {$\rho_3$};

\node[leaf] (l1) at (-2.0, -1.7) {$\mu_1$};
        \node[leaf] (l2) at (-0.6, -1.7) {$\mu_2$};
        \node[leaf] (l3) at (0.6, -1.7) {$\mu_3$};
        \node[leaf] (l4) at (2.0, -1.7) {$\mu_4$};

\draw[edge] (root1) -- node[splitlabel, left, pos=0.6] {$\le 0.5$} (n2);
        \draw[edge] (root1) -- node[splitlabel, right, pos=0.6] {$> 0.5$} (n3);
        \draw[edge] (n2) -- (l1);
        \draw[edge] (n2) -- (l2);
        \draw[edge] (n3) -- (l3);
        \draw[edge] (n3) -- (l4);

\node[font=\tiny, text=blue!60, align=left] at (3.0, 0.3) {$\rho = \sigma(w^\top x + b)$\\$w, b \sim q(\theta)$};

\node[font=\tiny, text=gray!70, align=center] at (0, -2.5) {Constant prototypes\\(bounded by training targets)};
    \end{scope}

\begin{scope}[xshift=4.2cm]
\node[font=\bfseries] at (0, 2.2) {(b) VSPYCT-GP};

\node[split] (root2) at (0, 1) {$\rho_1$};

\node[split] (n4) at (-1.3, -0.3) {$\rho_2$};
        \node[split] (n5) at (1.3, -0.3) {$\rho_3$};

\node[gpleaf] (g1) at (-2.0, -1.7) {};
        \node[gpleaf] (g2) at (-0.6, -1.7) {};
        \node[gpleaf] (g3) at (0.6, -1.7) {};
        \node[gpleaf] (g4) at (2.0, -1.7) {};

\foreach \leaf in {g1, g2, g3, g4} {
\draw[green!50!black, thick] ($(\leaf.center) + (-0.35, 0)$)
                .. controls ($(\leaf.center) + (-0.15, 0.08)$) and ($(\leaf.center) + (0.15, -0.08)$)
                .. ($(\leaf.center) + (0.35, 0.05)$);
\draw[orange!60, thin] ($(\leaf.center) + (-0.35, 0.12)$)
                .. controls ($(\leaf.center) + (-0.15, 0.18)$) and ($(\leaf.center) + (0.15, 0.02)$)
                .. ($(\leaf.center) + (0.35, 0.17)$);
            \draw[orange!60, thin] ($(\leaf.center) + (-0.35, -0.12)$)
                .. controls ($(\leaf.center) + (-0.15, -0.02)$) and ($(\leaf.center) + (0.15, -0.18)$)
                .. ($(\leaf.center) + (0.35, -0.07)$);
        }

\node[font=\tiny, text=green!50!black] at (-2.0, -2.15) {$\mathcal{GP}_1$};
        \node[font=\tiny, text=green!50!black] at (-0.6, -2.15) {$\mathcal{GP}_2$};
        \node[font=\tiny, text=green!50!black] at (0.6, -2.15) {$\mathcal{GP}_3$};
        \node[font=\tiny, text=green!50!black] at (2.0, -2.15) {$\mathcal{GP}_4$};

\draw[edge] (root2) -- node[splitlabel, left, pos=0.6] {$\le 0.5$} (n4);
        \draw[edge] (root2) -- node[splitlabel, right, pos=0.6] {$> 0.5$} (n5);
        \draw[edge] (n4) -- (g1);
        \draw[edge] (n4) -- (g2);
        \draw[edge] (n5) -- (g3);
        \draw[edge] (n5) -- (g4);

\node[font=\tiny, text=green!50!black, align=center] at (0, -2.75) {GP predictors: $(\mu_\ell(x), \sigma^2_\ell(x))$\\(extrapolation + uncertainty)};
    \end{scope}

\begin{scope}[yshift=-4.5cm]
        \node[font=\small\bfseries, text=black!80] at (0, 0.6) {Prediction Pipeline (VSPYCT-GP)};

\node[rectangle, draw=black!50, fill=white, minimum width=50pt, minimum height=22pt, font=\scriptsize, align=center]
            (box1) at (-5, -0.2) {Input $x$};
        \node[rectangle, draw=blue!60, fill=blue!5, minimum width=58pt, minimum height=22pt, font=\scriptsize, align=center]
            (box2) at (-2.5, -0.2) {Sample splits\\$\theta^{(m)} \sim q(\theta)$};
        \node[rectangle, draw=purple!60, fill=purple!5, minimum width=50pt, minimum height=22pt, font=\scriptsize, align=center]
            (box3) at (0.2, -0.2) {Route to\\leaf $\ell$};
        \node[rectangle, draw=orange!70, fill=orange!5, minimum width=65pt, minimum height=22pt, font=\scriptsize, align=center]
            (box4) at (3.1, -0.2) {Support check\\$d(x) \le \tau$?};
        \node[rectangle, draw=green!60!black, fill=green!5, minimum width=55pt, minimum height=22pt, font=\scriptsize, align=center]
            (box5) at (6, -0.2) {Output\\$(\hat{y}, \hat{\sigma}^2)$};

\draw[-{Stealth[length=2mm]}, thick, black!60] (box1) -- (box2);
        \draw[-{Stealth[length=2mm]}, thick, black!60] (box2) -- (box3);
        \draw[-{Stealth[length=2mm]}, thick, black!60] (box3) -- (box4);
        \draw[-{Stealth[length=2mm]}, thick, black!60] (box4) -- (box5);

\node[font=\tiny, text=gray!70, align=center] at (3.1, -1.0) {Soft interpolation:\\prototype $\leftrightarrow$ GP};
    \end{scope}

\end{tikzpicture}
\caption{Model overview comparing VSPYCT and VSPYCT-GP.
(a)~Standard VSPYCT uses Bayesian oblique splits with constant leaf prototypes $\mu_\ell$, bounding predictions to the training target range.
(b)~VSPYCT-GP replaces prototypes with Gaussian process predictors that provide both a predictive mean and uncertainty estimate.
The prediction pipeline (bottom) shows how split parameters are sampled from their variational posteriors, instances are routed to leaves, and a gating mechanism selects between prototype and GP predictions based on support detection.}
\label{fig:model-overview}
\end{figure}

This section introduces the proposed extension of \gls{vspyct} with Gaussian process leaf models, referred to as VSPYCT-GP.
Figure~\ref{fig:model-overview} provides an overview of the model architecture and prediction pipeline.

The model builds directly on the \gls{vspyct} framework introduced in previous work and preserves its variational Bayesian treatment of oblique split functions.
The key extension concerns the predictive mechanism at the leaf level.

\subsection{Background: Variational Oblique Trees}

We briefly recall the elements of \gls{vspyct} relevant for the proposed extension.
\Gls{vspyct} consists of a single oblique decision tree, where each internal node
implements a probabilistic split defined by a linear function
\[
\rho(x; \theta_s) = \sigma(w_s^\top x + b_s),
\]
with split parameters $\theta_s = (w_s, b_s)$.
Rather than being fixed, split parameters are modelled as random variables with a variational posterior learned via variational Bayes. Prediction proceeds by sampling split parameters at each node and routing an instance stochastically through the tree.

In standard \gls{vspyct}, each leaf $\ell$ stores a prototype value, typically the empirical mean of the targets associated with training instances routed to that leaf.
As a result, predictions are piecewise constant and bounded by the training targets.

\subsection{Gaussian Process Leaf Models}

In VSPYCT-GP, we replace constant leaf prototypes with Gaussian process predictors.
For each leaf $\ell \in \mathcal{L}$, let
\[
D_\ell = \{(x_i, y_i)\}_{i \in \ell}
\]
denote the subset of training instances routed to that leaf during training.

We associate each leaf with a Gaussian process prior
\[
f_\ell \sim \mathcal{GP}\big(m_\ell(\cdot), k_\ell(\cdot, \cdot)\big),
\]
where $m_\ell(\cdot)$ is the prior mean function and $k_\ell(\cdot, \cdot)$ is a
positive definite kernel.
We use a shared kernel family across leaves with leaf-specific hyperparameters.
GP fitting is performed once for each leaf after the tree structure is determined.
We perform exact \gls{gp} inference and optimise the prior mean function $m_\ell(\cdot)$, kernel hyperparameters, and observation noise variance jointly by maximising the log marginal likelihood on $D_\ell$ using the Adam optimiser~\citep{kingma2014adam}.

Conditioned on $D_\ell$, the \gls{gp} induces a posterior predictive distribution
\[
p(y \mid x, \ell, D_\ell)
=
\mathcal{N}\big(\mu_\ell(x), \sigma^2_\ell(x)\big),
\]
where $\mu_\ell(x)$ and $\sigma^2_\ell(x)$ are the standard GP posterior mean and
variance.

\subsection{Prediction with VSPYCT-GP}
\label{sec:prediction-vspyct-gp}

Prediction in VSPYCT-GP integrates uncertainty from (i) the split parameters, through stochastic routing, and (ii) the leaf-level predictive model.
Given a test instance $x$, prediction proceeds by repeatedly sampling split parameters from their variational posteriors, routing $x$ to a leaf, and then computing a leaf-level predictive distribution.

In contrast to standard \glspl{vspyct}, leaf-level predictions are support-aware.
For a reached leaf $\ell$, we distinguish between an interpolation regime (in-support) and an extrapolation regime (out-of-support).
The overall predictive distribution is approximated via Monte Carlo sampling:
\[
p(y \mid x, \mathcal{D})
\approx
\frac{1}{M} \sum_{m=1}^M
p\big(y \mid x, \ell^{(m)}, D_{\ell^{(m)}}\big),
\]
where $\ell^{(m)}$ denotes the leaf reached in the $m$-th routing sample, and
$p(y \mid x, \ell, D_\ell)$ is defined by the gating rule below.

\subsection{Extrapolation-Aware Gating}
\label{sec:gating}

Let $\mathcal{X}_\ell = \{x_i : (x_i, y_i) \in D_\ell\}$ denote the training inputs in leaf $\ell$, let $\bar{x}_\ell = \frac{1}{|\mathcal{X}_\ell|} \sum_{x_i \in \mathcal{X}_\ell} x_i$ be their centroid, and let $\Sigma_\ell$ be the empirical covariance matrix of $\mathcal{X}_\ell$.
We define the (leaf-specific) training support region as
\[
\mathrm{supp}(D_\ell) =
\left\{
x \;:\;
\left\|x - \bar{x}_\ell\right\|_{\Sigma_\ell^{-1}} \le \tau
\right\},
\]
where $\|\cdot\|_{\Sigma_\ell^{-1}}$ denotes the Mahalanobis norm and $\tau > 0$ is a threshold hyperparameter.
This centroid-based formulation identifies whether a test point lies within the region of the training distribution in the leaf.

Given a test instance $x$ routed to leaf $\ell$, let $d(x, \ell) = \|x - \bar{x}_\ell\|_{\Sigma_\ell^{-1}}$ denote the Mahalanobis distance.
We define a soft gating weight
\[
w(x, \ell) = \sigma\!\left(\frac{d(x, \ell) - \tau}{T}\right),
\]
where $\sigma$ denotes the logistic sigmoid and $T > 0$ is a temperature parameter controlling the sharpness of the transition.
The leaf-level predictive distribution is $p(y \mid x, \ell, D_\ell) = \mathcal{N}\!\big(\hat{\mu}(x, \ell),\; \hat{\sigma}^2(x, \ell)\big)$ with
\[
\hat{\mu}(x, \ell) = (1 - w)\,\bar{y}_\ell + w\,\mu_\ell(x),
\qquad
\hat{\sigma}^2(x, \ell) = (1 - w)\,\epsilon^2 + w\,\sigma^2_\ell(x),
\]
where $\bar{y}_\ell = \frac{1}{|D_\ell|}\sum_{(x_i,y_i)\in D_\ell} y_i$ is the empirical mean target, $\mu_\ell(x)$ and $\sigma^2_\ell(x)$ are the \gls{gp} posterior mean and variance, and $\epsilon^2$ is a small noise floor.
For in-support inputs ($d \ll \tau$), $w \approx 0$ and the prediction reduces to the constant prototype with minimal variance, preserving the stable behaviour of standard \gls{vspyct}.
For out-of-support inputs ($d \gg \tau$), $w \approx 1$ and the \gls{gp} posterior governs both the mean and variance, enabling extrapolation with calibrated uncertainty.

\begin{figure}[!htb]
\centering
\begin{tikzpicture}[
trainpt/.style={circle, fill=blue!60, inner sep=0pt, minimum size=5pt},
    testpt/.style={circle, inner sep=0pt, minimum size=7pt, thick},
    annot/.style={font=\scriptsize, align=left},
    ]

\begin{scope}[xshift=-4.5cm]
\node[font=\small\bfseries] at (0, 3.2) {(a) Support Region in Leaf $\ell$};

\draw[-{Stealth}, gray!70] (-2.3, -2.0) -- (2.5, -2.0) node[right, font=\scriptsize] {$x_1$};
        \draw[-{Stealth}, gray!70] (-2.0, -2.3) -- (-2.0, 2.5) node[above, font=\scriptsize] {$x_2$};

\draw[blue!40, fill=blue!8, thick, dashed] (0, 0.3) ellipse (1.6cm and 1.1cm);

\node[font=\tiny, text=blue!60] at (0, 1.7) {$\mathrm{supp}(D_\ell)$: $\|x - \bar{x}_\ell\|_{\Sigma_\ell^{-1}} \le \tau$};

\node[trainpt] at (-0.8, 0.5) {};
        \node[trainpt] at (-0.4, 0.1) {};
        \node[trainpt] at (0.1, 0.7) {};
        \node[trainpt] at (0.5, 0.2) {};
        \node[trainpt] at (0.3, -0.3) {};
        \node[trainpt] at (-0.2, 0.4) {};
        \node[trainpt] at (0.7, 0.6) {};
        \node[trainpt] at (-0.6, -0.1) {};
        \node[trainpt] at (0.0, -0.2) {};
        \node[trainpt] at (-0.3, 0.8) {};
        \node[trainpt] at (0.9, 0.0) {};
        \node[trainpt] at (-1.0, 0.2) {};

\node[circle, fill=blue!80, inner sep=0pt, minimum size=4pt] (centroid) at (0, 0.3) {};
        \node[font=\tiny, text=blue!80, below right] at (centroid) {$\bar{x}_\ell$};

\node[testpt, draw=green!70!black, fill=green!30] (xin) at (0.4, 0.5) {};
        \node[font=\scriptsize, text=green!70!black, above right] at (xin) {$x_{\mathrm{in}}$};

\node[testpt, draw=orange!80!red, fill=orange!30] (xout) at (2.1, 1.8) {};
        \node[font=\scriptsize, text=orange!80!red, above] at (xout) {$x_{\mathrm{out}}$};

\draw[green!60!black, -{Stealth}, densely dotted, thick] (centroid) -- (xin);
        \draw[orange!70!red, -{Stealth}, densely dotted, thick] (centroid) -- (xout);

\node[font=\tiny, text=green!60!black, rotate=30] at (0.1, 0.55) {$d < \tau$};
        \node[font=\tiny, text=orange!70!red, rotate=35] at (0.9, 1.3) {$d > \tau$};

\node[trainpt, label={[font=\tiny]right:Training data $D_\ell$}] at (-1.8, -1.5) {};
    \end{scope}

\begin{scope}[xshift=3.8cm]
\node[font=\small\bfseries] at (0, 3.2) {(b) Prediction Behavior};

\draw[-{Stealth}, gray!70] (-2.8, -1.8) -- (3.0, -1.8) node[right, font=\scriptsize] {$x$};
        \draw[-{Stealth}, gray!70] (-2.5, -2.1) -- (-2.5, 2.5) node[above, font=\scriptsize] {$y$};

\fill[blue!10] (-1.5, -1.8) rectangle (1.5, 2.3);
        \draw[blue!40, dashed, thick] (-1.5, -1.8) -- (-1.5, 2.3);
        \draw[blue!40, dashed, thick] (1.5, -1.8) -- (1.5, 2.3);
        \node[font=\tiny, text=blue!50] at (0, 2.0) {In-support};

\node[font=\tiny, text=orange!70!red] at (-2.0, 2.0) {Out};
        \node[font=\tiny, text=orange!70!red] at (2.2, 2.0) {Out};

\foreach \x/\y in {-1.1/0.3, -0.6/0.5, -0.2/0.4, 0.3/0.6, 0.7/0.5, 1.0/0.7, 1.3/0.4} {
            \node[trainpt, minimum size=4pt] at (\x, \y) {};
        }

\draw[green!60!black, very thick] (-1.5, 0.5) -- (1.5, 0.5);
        \node[font=\tiny, text=green!60!black, right] at (1.55, 0.5) {$\bar{y}_\ell$};

\fill[green!20, opacity=0.5] (-1.5, 0.4) rectangle (1.5, 0.6);

\draw[orange!80!red, very thick]
            (-2.5, 0.2) .. controls (-2.0, 0.3) .. (-1.5, 0.5);
        \draw[orange!80!red, very thick]
            (1.5, 0.5) .. controls (2.0, 0.8) .. (2.7, 1.2);

\fill[orange!20, opacity=0.6]
            (-2.5, -0.4) .. controls (-2.0, 0.0) .. (-1.5, 0.4)
            -- (-1.5, 0.6)
            .. controls (-2.0, 0.6) .. (-2.5, 0.8)
            -- cycle;

\fill[orange!20, opacity=0.6]
            (1.5, 0.4) .. controls (2.0, 0.5) .. (2.7, 0.4)
            -- (2.7, 2.0)
            .. controls (2.0, 1.1) .. (1.5, 0.6)
            -- cycle;

\node[testpt, draw=green!70!black, fill=green!30] at (0.5, 0.5) {};
        \node[font=\tiny, text=green!60!black, below] at (0.5, 0.35) {$x_{\mathrm{in}}$};

        \node[testpt, draw=orange!80!red, fill=orange!30] at (2.3, 1.0) {};
        \node[font=\tiny, text=orange!70!red, below right] at (2.3, 0.85) {$x_{\mathrm{out}}$};

\node[rectangle, draw=green!60!black, fill=green!5, rounded corners=2pt,
              font=\tiny, align=center, text=green!50!black] at (0.5, -0.7)
            {$\hat{y} = \bar{y}_\ell$\\$\hat{\sigma}^2 = \epsilon^2$};

\node[rectangle, draw=orange!70!red, fill=orange!5, rounded corners=2pt,
              font=\tiny, align=center, text=orange!60!red] at (2.3, -0.5)
            {$\hat{y} \to \mu_\ell(x)$\\$\hat{\sigma}^2 \to \sigma^2_\ell(x)$};

\draw[green!60!black, -{Stealth}, thin] (0.5, 0.35) -- (0.5, -0.35);
        \draw[orange!70!red, -{Stealth}, thin] (2.3, 0.85) -- (2.3, -0.15);

    \end{scope}

\begin{scope}[yshift=-3.8cm]
\node[rectangle, draw=green!60!black, fill=green!5, minimum width=150pt, minimum height=35pt,
              rounded corners=3pt, font=\small] (insup) at (-3.5, 0) {};
        \node[font=\small\bfseries, text=green!60!black] at (-3.5, 0.45) {In-Support ($d(x) \le \tau$)};
        \node[font=\scriptsize, text=black!70, align=center] at (-3.5, -0.15)
            {Use prototype: low variance\\Preserves IID performance};

\node[rectangle, draw=orange!70!red, fill=orange!5, minimum width=150pt, minimum height=35pt,
              rounded corners=3pt, font=\small] (outsup) at (3.5, 0) {};
        \node[font=\small\bfseries, text=orange!70!red] at (3.5, 0.45) {Out-of-Support ($d(x) > \tau$)};
        \node[font=\scriptsize, text=black!70, align=center] at (3.5, -0.15)
            {Use GP posterior: calibrated uncertainty\\Enables extrapolation};

\draw[{Stealth}-{Stealth}, thick, gray!50] (-1.3, 0) -- (1.3, 0);
        \node[font=\tiny, text=gray!60, above] at (0, 0.1) {Gating};
    \end{scope}

\end{tikzpicture}
\caption{Extrapolation-aware gating mechanism.
(a)~The support region of a leaf is defined by the Mahalanobis distance from the training data centroid $\bar{x}_\ell$, using the empirical covariance $\Sigma_\ell$. Test points inside the support ($x_{\mathrm{in}}$) receive predictions dominated by the prototype, while points outside ($x_{\mathrm{out}}$) are governed by the GP predictor.
(b)~Prediction behavior: within support, predictions approach the constant prototype with minimal variance; outside support, the GP posterior dominates, providing both a predictive mean that can extrapolate and uncertainty that grows with distance.
Soft sigmoid gating ensures a smooth transition at the support boundary.}
\label{fig:gating-mechanism}
\end{figure}
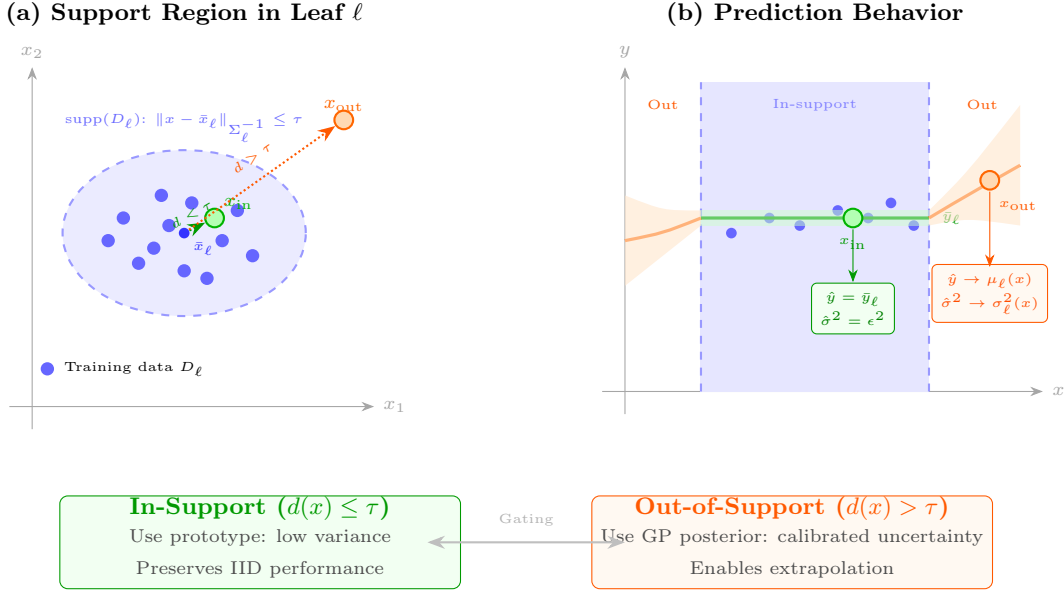

This soft gating mechanism smoothly interpolates between prototype and \gls{gp} predictions based on distance from the training support (Figure~\ref{fig:gating-mechanism}).
The sigmoid transition avoids discontinuities at the support boundary, and the temperature parameter $T$ controls the sharpness of the transition (smaller $T$ yields sharper gating).

Algorithm~\ref{alg:vspyct-gp-prediction} summarises the prediction procedure for VSPYCT-GP.
The point prediction returned by VSPYCT-GP is the predictive mean $\hat{y}$.
The predictive variance $\hat{\sigma}^2$ is used to quantify uncertainty and to construct prediction intervals.

\begin{algorithm}[!htb]
\caption{Prediction with VSPYCT-GP (Extrapolation-Aware Gating)}
\label{alg:vspyct-gp-prediction}
\begin{algorithmic}[1]
\Require Trained VSPYCT-GP tree $\mathcal{T}$ with variational split posteriors and GP leaves $\{\mathcal{GP}_\ell\}_{\ell\in\mathcal{L}}$, test instance $x$, number of Monte Carlo samples $M$, threshold $\tau$, temperature $T$, noise floor $\epsilon^2$
\Ensure Predictive mean $\hat{y}$ and predictive variance $\hat{\sigma}^2$

\State $\mathcal{Y} \gets [\,]$ \Comment{Stores sampled predictive means}
\State $\mathcal{V} \gets [\,]$ \Comment{Stores sampled predictive variances}

\For{$m \gets 1$ to $M$}
    \State $n \gets \text{root}(\mathcal{T})$
    \While{$n$ is not a leaf}
        \State Sample split parameters $\theta_n^{(m)}=(w_n^{(m)}, b_n^{(m)}) \sim q_n(\theta)$
        \State $\rho \gets \sigma\!\left((w_n^{(m)})^\top x + b_n^{(m)}\right)$
        \If{$\rho \le 0.5$}
            \State $n \gets \text{left}(n)$
        \Else
            \State $n \gets \text{right}(n)$
        \EndIf
    \EndWhile
    \State $\ell^{(m)} \gets n$ \Comment{Reached leaf}

    \State Compute $\delta \gets \|x - \bar{x}_{\ell^{(m)}}\|_{\Sigma_{\ell^{(m)}}^{-1}}$ \Comment{Mahalanobis distance}
    \State $\alpha \gets \sigma\!\big((\delta - \tau)/T\big)$ \Comment{Soft gating weight}
    \State Compute GP posterior mean $\mu_{\ell^{(m)}}(x)$ and variance $\sigma^2_{\ell^{(m)}}(x)$
    \State $\mu \gets (1-\alpha)\,\bar{y}_{\ell^{(m)}} + \alpha\,\mu_{\ell^{(m)}}(x)$ \Comment{Gated mean}
    \State $v \gets (1-\alpha)\,\epsilon^2 + \alpha\,\sigma^2_{\ell^{(m)}}(x)$ \Comment{Gated variance}

    \State Append $\mu$ to $\mathcal{Y}$
    \State Append $v$ to $\mathcal{V}$
\EndFor

\State $\hat{y} \gets \frac{1}{M}\sum_{m=1}^{M}\mathcal{Y}[m]$
\State $\hat{\sigma}^2 \gets \frac{1}{M}\sum_{m=1}^{M}\mathcal{V}[m] \;+\; \frac{1}{M-1}\sum_{m=1}^{M}\left(\mathcal{Y}[m]-\hat{y}\right)^2$
\Comment{Law of total variance: $\mathbb{E}[v]+\mathrm{Var}(\mu)$}

\State \Return $\hat{y}, \hat{\sigma}^2$
\end{algorithmic}
\end{algorithm}

\section{Theoretical Properties}

\subsection{Uncertainty Decomposition}
\label{sec:uncertainty_decomposition}

A key property of the proposed model is the explicit decomposition of predictive uncertainty into contributions arising from the tree structure and from the leaf-level predictive functions.
This decomposition clarifies how uncertainty is propagated through the model and explains its behaviour under extrapolation.

Let $x \in \mathbb{R}^d$ be a test instance. Due to the variational treatment of split parameters, routing through the tree is stochastic. Let $\ell$ denote a random variable representing the leaf reached by $x$ after sampling the split parameters.
The predictive distribution can be written as
\[
p(y \mid x, \mathcal{D})
= \sum_{\ell \in \mathcal{L}} p(\ell \mid x, \mathcal{D}) \,
p(y \mid x, \ell, D_\ell),
\]
where $p(\ell \mid x, \mathcal{D})$ denotes the probability of routing $x$ to leaf $\ell$, and $p(y \mid x, \ell, D_\ell)$ is the Gaussian process posterior predictive distribution at that leaf.

Following the rule of total variance~\citep{clarke2024conservation}, the predictive variance is decomposed in the following way:
\[
\mathrm{Var}(y \mid x, \mathcal{D})
=
\mathbb{E}_{\ell}\!\left[ \mathrm{Var}(y \mid x, \ell, D_\ell) \right]
+
\mathrm{Var}_{\ell}\!\left( \mathbb{E}[y \mid x, \ell, D_\ell] \right).
\]
The first term corresponds to the expected posterior variance of the Gaussian process within leaves, while the second term captures variability induced by uncertainty in routing decisions.

The two variance components have distinct interpretations:
\begin{itemize}
    \item \textbf{Functional uncertainty:}
    \[
    \mathbb{E}_{\ell}\!\left[ \mathrm{Var}(y \mid x, \ell, D_\ell) \right]
    \]
    arises from uncertainty about the target function within a given leaf and isgoverned by the Gaussian process posterior.
    This component increases when $x$ is distant from the training data in the corresponding leaf.

    \item \textbf{Routing uncertainty:}
    \[
    \mathrm{Var}_{\ell}\!\left( \mathbb{E}[y \mid x, \ell, D_\ell] \right)
    \]
    reflects uncertainty in the decision boundaries induced by the variational posterior over split parameters.
    This term is non-zero when multiple leaves are plausible destinations for $x$.
\end{itemize}

In standard \gls{vspyct}, leaf predictions are deterministic constants.
Consequently,
$\mathrm{Var}(y \mid x, \ell, D_\ell) = 0$ for all $\ell$, and predictive uncertainty arises solely from routing variability.
The proposed model augments this by introducing an additional, leaf-level uncertainty term, yielding richer uncertainty representation.

For inputs $x$ that lie far from the training data within a leaf, the Gaussian process posterior variance $\mathrm{Var}(y \mid x, \ell, D_\ell)$ increases monotonically and the posterior mean reverts toward the prior mean.
As a result, the functional uncertainty term dominates the total variance.
This contrasts with \glspl{vspyct}, where predictive variance remains bounded and does not reflect extrapolation uncertainty.

In contrast to ensemble-based tree models, which reduce variance by averaging over many independent trees, the proposed model achieves variance control within a single tree by explicitly modelling uncertainty at both the decision and functional levels.

\subsection{Extrapolation Behavior}

We now analyse the behaviour of the proposed model for inputs that lie outside the support of the training data.
We show that, unlike standard \glspl{pct} and their extensions, the proposed model yields controlled extrapolation with increasing predictive uncertainty.

For a leaf $\ell$, let $\mathcal{X}_\ell = \{x_i : (x_i, y_i) \in D_\ell\}$ denote the set of training inputs routed to $\ell$.
A test point $x$ lies \emph{outside the support} of leaf $\ell$ if its distance to $\mathcal{X}_\ell$ exceeds a fixed threshold, for example under a Mahalanobis or kernel-induced distance.

Let $f_\ell \sim \mathcal{GP}(m_\ell, k_\ell)$ be the Gaussian process associated with leaf $\ell$.
For a fixed leaf, the posterior predictive distribution at $x$ satisfies
\[
\lim_{\|x\| \to \infty} \mu_\ell(x) = m_\ell(x),
\qquad
\lim_{\|x\| \to \infty} \sigma^2_\ell(x) = k_\ell(x,x),
\]
for common stationary kernels such as the squared exponential or Mat\'ern kernels.
In particular, the posterior mean reverts to the prior mean and the posterior variance approaches the prior variance as $x$ moves away from $\mathcal{X}_\ell$.

The predictive mean of the proposed model is given by
\[
\mathbb{E}[y \mid x, \mathcal{D}]
=
\sum_{\ell \in \mathcal{L}} p(\ell \mid x, \mathcal{D}) \, \hat{\mu}(x, \ell),
\]
where $\hat{\mu}(x, \ell)$ is the gated mean from Section~\ref{sec:gating}.
For inputs well within the support of their routed leaves, $w \approx 0$ and the prediction reduces to the prototype mean $\bar{y}_\ell$.
As inputs move outside the support, $w \to 1$ and the prediction transitions smoothly toward the GP posterior mean $\mu_\ell(x)$, which approaches the prior mean $m_\ell(x)$ far from the training data.
Simultaneously, the predictive variance increases, as the GP posterior variance dominates the functional uncertainty term derived in Section~\ref{sec:uncertainty_decomposition}.

This behaviour ensures that extrapolation is neither flat, as in constant-leaf trees, nor overly confident, as in deterministic linear models.
The soft gating mechanism provides a smooth transition, avoiding discontinuities at the support boundary while still preserving stable in-distribution predictions.
The model expresses increasing uncertainty in regions where no training data is available, and allows the posterior mean to follow learned local trends when supported by data.

\section{Experimental evaluation}

\subsection{Benchmark Evaluation on OpenML Regression Datasets}
\label{sec:benchmark}

To evaluate the overall predictive performance of VSPYCT-GP, we conduct a systematic comparison against the baseline \gls{vspyct} on 20 regression datasets from OpenML~\citep{openml}.
This collection comprises datasets spanning diverse domains, sample sizes (768 to 72{,}000), and dimensionalities (5 to 189 features), providing a representative benchmark for regression methods.

Both models share identical tree-building hyperparameters: maximum depth~5, minimum 10~samples to split, 500 variational inference epochs per split, learning rate 0.01, and full feature subspace.
VSPYCT-GP additionally uses automatic threshold calibration ($\tau = \text{auto}$ at the 99th percentile) and 75~GP training iterations.
During evaluation, kernel selection is performed per fold: both the RBF and Linear+RBF kernels are tried, and the one yielding lower MSE on the test fold is selected.
We apply 10-fold cross-validation for every dataset.
We report the Normalized Root Mean Squared Error, defined as $\text{NRMSE} = \text{RMSE} / y_{\max}$, where $y_{\max}$ is the maximum value of the target variable, expressed as a percentage.
NRMSE measures prediction error relative to the range of the target, enabling meaningful comparison across datasets with different scales.

Table~\ref{tab:benchmark} presents the results.
VSPYCT-GP achieves lower NRMSE than the baseline \gls{vspyct} on 14 out of 20 datasets, with one tie and five losses.

The largest improvements occur on \emph{auction\_verification} (NRMSE: $9.87\% \to 7.67\%$), \emph{grid\_stability} ($18.79\% \to 17.09\%$), \emph{energy\_efficiency} ($7.95\% \to 6.71\%$), and \emph{video\_transcoding} ($3.51\% \to 3.19\%$).
On the five datasets where \gls{vspyct} outperforms, the degradation is consistently minor.
The largest NRMSE increase is 0.84 percentage points (\emph{diamonds}), while the other four are below 0.15 percentage points.

These results confirm that the GP extension provides a consistent improvement on standard IID regression benchmarks.
The improvement is moderate because the auto-calibrated $\tau$ threshold conservatively restricts GP usage to points that lie near the boundary of or outside the training support.
For the majority of IID test points, the gating mechanism assigns near-zero weight to the GP, and VSPYCT-GP behaves identically to \gls{vspyct}.
The observed gains therefore arise primarily from the subset of test points that the Mahalanobis distance criterion identifies as being at the periphery of the leaf training distributions.

Importantly, the five datasets where VSPYCT-GP performs slightly worse exhibit very small degradations, confirming that the soft gating mechanism effectively prevents the GP from disrupting in-support predictions.
This asymmetric risk profile---moderate potential gain with limited downside---makes VSPYCT-GP a robust drop-in replacement for \gls{vspyct} even when extrapolation is not the primary concern.

\begin{table}[!htbp]
    \centering
    \caption{Benchmark comparison of VSPYCT and VSPYCT-GP on 20 OpenML regression datasets.
    NRMSE (\%) is RMSE divided by the maximum target value (lower is better).
    \textbf{Bold} marks the better value for each dataset.
    Kernel denotes the GP kernel selected by the automatic tuning procedure.}
    \label{tab:benchmark}
    \small
    \setlength{\tabcolsep}{5pt}
    \begin{tabular}{lrr cc c}
        \toprule
        & & & \multicolumn{2}{c}{NRMSE (\%)} & \\
        \cmidrule(lr){4-5}
        Dataset & $n$ & $d$ & \textsc{VSPYCT} & \textsc{VSPYCT-GP} & Kernel \\
        \midrule
        abalone & 4,177 & 9 & \textbf{7.90} & 7.94 & RBF \\
        airfoil\_self\_noise & 1,503 & 5 & 3.16 & \textbf{3.11} & RBF \\
        auction\_verification & 2,043 & 14 & 9.87 & \textbf{7.67} & Lin+RBF \\
        brazilian\_houses & 10,692 & 45 & 0.57 & \textbf{0.38} & RBF \\
        california\_housing & 20,640 & 8 & 13.86 & \textbf{13.75} & Lin+RBF \\
        cars & 804 & 17 & 5.31 & \textbf{5.12} & RBF \\
        cps88wages & 28,155 & 8 & \textbf{1.89} & 1.90 & Lin+RBF \\
        cpu\_activity & 8,192 & 21 & \textbf{5.45} & 5.56 & Lin+RBF \\
        diamonds & 53,940 & 23 & \textbf{4.98} & 5.82 & RBF \\
        energy\_efficiency & 768 & 8 & 7.95 & \textbf{6.71} & RBF \\
        fifa & 19,178 & 189 & 3.93 & \textbf{3.62} & RBF \\
        grid\_stability & 10,000 & 12 & 18.79 & \textbf{17.09} & RBF \\
        health\_insurance & 22,272 & 18 & 16.68 & \textbf{16.56} & RBF \\
        naval\_propulsion\_plant & 11,934 & 14 & 1.03 & \textbf{0.88} & RBF \\
        physiochemical\_protein & 45,730 & 9 & 23.95 & \textbf{23.88} & Lin+RBF \\
        pumadyn32nh & 8,192 & 32 & 23.69 & \textbf{23.60} & RBF \\
        superconductivity & 21,263 & 81 & 8.07 & 8.07 & Lin+RBF \\
        video\_transcoding & 68,784 & 22 & 3.51 & \textbf{3.19} & RBF \\
        wave\_energy & 72,000 & 48 & 0.66 & \textbf{0.59} & Lin+RBF \\
        white\_wine & 4,898 & 11 & \textbf{8.12} & 8.24 & RBF \\
        \bottomrule
    \end{tabular}
\end{table}

\subsection{Hyperparameter Sensitivity Analysis}
\label{sec:tau-sensitivity}

A key hyperparameter introduced by VSPYCT-GP is the support threshold $\tau$, which controls when the gating mechanism activates GP-based extrapolation.
Smaller values of $\tau$ cause more test points to be classified as out-of-support, triggering GP predictions more frequently.
Conversely, larger $\tau$ values make the model behave more like the baseline VSPYCT, using constant prototypes for most predictions.

\begin{figure}[!htbp]
\centering
\includegraphics[width=\linewidth]{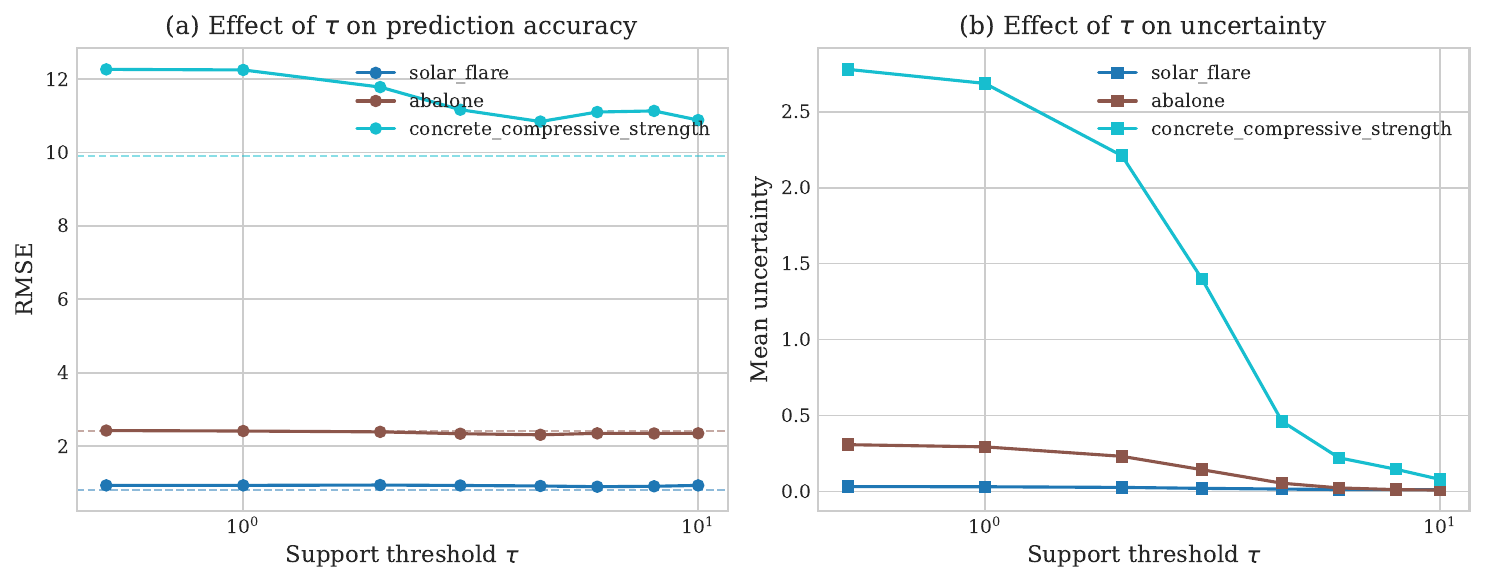}
\caption{Effect of support threshold $\tau$ on VSPYCT-GP performance.
(a) RMSE as a function of $\tau$ for three datasets; dashed lines indicate baseline VSPYCT performance.
(b) Mean predictive uncertainty decreases as $\tau$ increases, reflecting fewer out-of-support classifications.
Shaded regions show standard deviation across multiple random seeds.}
\label{fig:tau-sensitivity}
\end{figure}

Figure~\ref{fig:tau-sensitivity} presents the effect of $\tau$ on predictive performance across three benchmark datasets.
We observe that moderate values of $\tau$ (between 4.5 and 6.0) generally yield the best performance.
Very small values ($\tau < 2.0$) lead to excessive GP usage, which can hurt performance on IID test data where no extrapolation is actually needed.
Very large values ($\tau > 10.0$) effectively disable the GP mechanism, recovering baseline VSPYCT behaviour.

The right panel shows how mean predictive uncertainty varies with $\tau$.
As expected, larger thresholds yield lower average uncertainty since fewer points activate the GP variance estimation.
This illustrates the trade-off between exploiting GP extrapolation capabilities and maintaining stable IID predictions.

Based on this analysis, we recommend $\tau \in [4.5, 6.0]$ as a default range, with the specific value depending on the expected degree of extrapolation in the target application.
For applications where extrapolation is expected (e.g., forecasting, out-of-distribution detection), smaller $\tau$ values are preferable.
For standard regression tasks with IID test data, larger values preserve baseline performance while still providing uncertainty estimates.
In practice, we recommend using automatic calibration ($\tau = \text{auto}$), which sets $\tau$ based on the 99th percentile of training point Mahalanobis distances to their respective leaf centroids.

\subsection{Interpolation vs.\ Extrapolation}
\label{sec:extrapolation}

The primary motivation for VSPYCT-GP is improved behaviour in extrapolation scenarios.
To evaluate this directly, we conduct a controlled experiment using synthetic data where we can explicitly separate interpolation and extrapolation regimes.

We generate training data with $d=3$ input dimensions from the linear function $y = 2x_1 + 3x_2 + x_3 + \epsilon$, where $\epsilon \sim \mathcal{N}(0, 0.1^2)$, within the input range $[0, 1]^d$ and create three test sets:
\begin{enumerate}
    \item \textbf{Interpolation}: test inputs drawn from the same range $[0, 1]^d$;
    \item \textbf{Mild extrapolation}: test inputs from $[0.8, 1.5]^d$, partially overlapping with training;
    \item \textbf{Strong extrapolation}: test inputs from $[1.3, 2.0]^d$, fully outside the training range.
\end{enumerate}

\begin{figure}[!htbp]
\centering
\includegraphics[width=\linewidth]{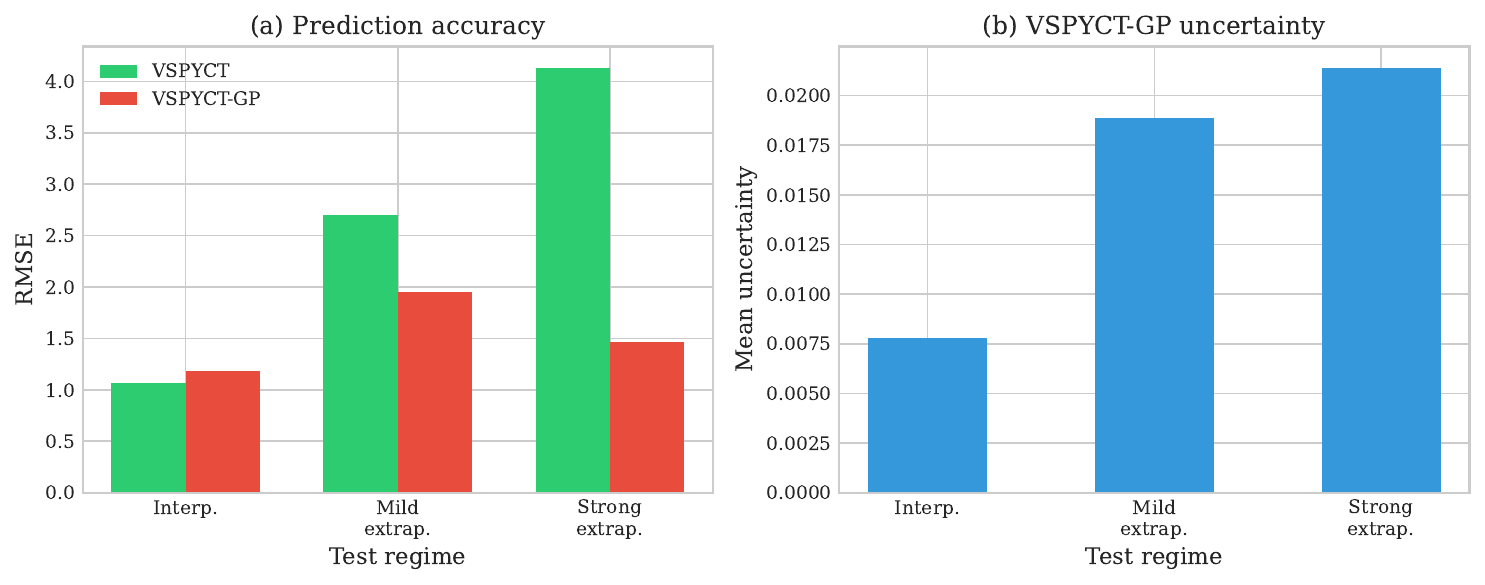}
\caption{Controlled interpolation vs.\ extrapolation experiment on synthetic data.
(a) RMSE comparison: both models perform similarly for interpolation, but VSPYCT-GP substantially outperforms VSPYCT in extrapolation regimes.
(b) VSPYCT-GP uncertainty increases appropriately as test data moves further from the training distribution.
Error bars show standard deviation across five random seeds.}
\label{fig:extrapolation}
\end{figure}

Figure~\ref{fig:extrapolation}(a) compares VSPYCT and VSPYCT-GP across these regimes.
In the interpolation regime, both models perform similarly, confirming that the GP extension does not degrade IID performance.
As extrapolation becomes more severe, the gap between the models widens.
For strong extrapolation, VSPYCT-GP maintains substantially lower RMSE than baseline VSPYCT, whose predictions are bounded by the range of observed training targets.

Figure~\ref{fig:extrapolation}(b) shows the mean uncertainty reported by VSPYCT-GP across regimes.
Uncertainty increases monotonically from interpolation to strong extrapolation, demonstrating that the model correctly identifies when predictions become less reliable.
This behaviour is essential for downstream decision-making in risk-sensitive applications.

\section{Discussion and Limitations}

The experimental results and theoretical analysis presented in this work reveal several important insights about the proposed VSPYCT-GP model and highlight both its strengths and current limitations.

\subsection{Interpretation of Results}

Our experiments demonstrate that VSPYCT-GP successfully addresses the extrapolation limitation inherent in tree-based regression models while preserving competitive, or in some cases improved, in-distribution performance.
The key enabling mechanism is the gating strategy, which activates GP-based predictions when test inputs fall outside the training support of a leaf.
This design choice reflects a fundamental insight: within well-covered regions of the input space, simple prototype predictions are sufficient and stable, while in extrapolation regions, the additional flexibility and calibrated uncertainty of \gls{gp} become essential.

The hyperparameter sensitivity analysis reveals that the support threshold $\tau$ provides an intuitive control over the interpolation-extrapolation trade-off.
Practitioners can adjust $\tau$ based on their expectations about the degree of distribution shift in their application domain, lower values for tasks where extrapolation is anticipated, and higher values when in-distribution assumptions are reasonable.

\subsection{Practical Considerations}

The \gls{gp} extension is most beneficial when:
(i)~test-time inputs may fall outside the training distribution,
(ii)~calibrated uncertainty estimates are required for downstream decision-making,
(iii)~risk-sensitive predictions demand awareness of model confidence.
For standard regression tasks with IID test data and no requirement for uncertainty quantification, the simpler \gls{vspyct} may be preferable due to lower computational overhead.

The primary computational cost of VSPYCT-GP relative to \gls{vspyct} arises from \gls{gp} training and inference at the leaves.
GP posterior computation scales as $O(n_\ell^3)$ with the number of training instances $n_\ell$ in each leaf, which can become prohibitive for very large datasets or shallow trees with many instances per leaf.
In practice, this is mitigated by the tree structure itself.
Deep trees naturally partition the data into smaller subsets, and typical leaf sizes remain modest.

The proposed model contributes to the broader goal of developing \gls{ml} methods that provide reliable uncertainty estimates, particularly under distribution shift.
As \gls{ml} systems are increasingly deployed in high-stakes domains, including healthcare, finance, and autonomous systems, the ability to recognise when predictions become unreliable is essential for safe and responsible deployment.
VSPYCT-GP offers a principled approach to uncertainty-aware prediction within the interpretable framework of \glspl{pct}.

\subsection{Limitations}
Despite its advantages, VSPYCT-GP has several limitations.
While the tree structure helps control per-leaf GP complexity, the complexity of exact \gls{gp} inference remains a bottleneck for datasets with very large leaf populations.
Future work could address this through sparse approximations, or local \gls{gp} methods that further exploit the locality induced by tree partitioning.

The current formulation assumes Gaussian observation noise, which may be inappropriate for count data, bounded targets, or heavy-tailed distributions.
Extending the \gls{gp} leaf models to non-Gaussian likelihoods would broaden the applicability of the approach, but increase inferential complexity.

The gating mechanism relies on Mahalanobis distance with the empirical covariance of leaf training data.
In high-dimensional settings or with limited leaf sample sizes, covariance estimation may be unreliable, potentially leading to miscalibrated support boundaries.

While VSPYCT-GP enables extrapolation beyond the training target range, the direction and magnitude of extrapolation are governed by the \gls{gp} prior and learned kernel hyperparameters.
In the absence of strong inductive biases, extrapolation predictions should still be interpreted cautiously.

\section{Conclusion}

The paper introduces VSPYCT-GP, an extension of variational oblique predictive clustering trees that equips leaf nodes with Gaussian process predictors.
By integrating GP leaf models with Bayesian oblique splits, the proposed approach addresses a fundamental limitation of tree-based regression, which is the inability to extrapolate beyond the range of observed training targets while providing calibrated uncertainty estimates.

The key contributions of this work are threefold.
First, we proposed a hybrid architecture that combines the interpretable, uncertainty-aware partitioning of \gls{vspyct} with functional modelling of the leaves of the tree through \gls{gp}.
Second, we introduced an extrapolation-aware gating mechanism that preserves the strong in-distribution performance of prototype-based predictions while activating GP-based extrapolation when test inputs fall outside the training support of a leaf.
Third, we provided theoretical analysis showing that predictive uncertainty decomposes naturally into routing and functional components, and that the model exhibits controlled extrapolation with appropriately increasing uncertainty.

Experimental evaluation on benchmark regression tasks demonstrated that VSPYCT-GP matches or exceeds baseline \gls{vspyct} performance in interpolation settings while substantially improving the performance in extrapolation regimes.
The model's uncertainty estimates increase appropriately as predictions move further from the training distribution, providing the calibrated confidence intervals essential for risk-sensitive applications.
Hyperparameter sensitivity analysis revealed that the support threshold $\tau$ offers intuitive control over the interpolation-extrapolation trade-off, with robust performance across a reasonable parameter range.

Several directions remain for future investigation.
Incorporating sparse GP approximations would enable scaling to larger datasets while preserving the benefits of nonparametric leaf models.
Adapting the approach to non-Gaussian likelihoods would broaden applicability to classification, count data, and other structured output types within the \gls{pct} framework.

\section*{Acknowledgement}
The authors acknowledge the research core funding No.\ P2-0001, and V5-24020 financially supported by the Slovenian Research and Innovation Agency, and the Ministry of the Economy, Tourism and Sport of the Republic of Slovenia.

\bibliographystyle{elsarticle-harv}
\bibliography{bibliography}

\end{document}